\documentclass[lettersize,journal]{IEEEtran}

\usepackage{amsmath,amsfonts}
\usepackage{algorithmic}
\usepackage{array}
\usepackage{textcomp}
\usepackage{stfloats}
\usepackage{url}
\usepackage{verbatim}
\usepackage{graphicx}
\def\BibTeX{{\rm B\kern-.05em{\sc i\kern-.025em b}\kern-.08em T\kern-.1667em\lower.7ex\hbox{E}\kern-.125emX}}
\usepackage{balance}

\usepackage{enumitem}
\usepackage{soul}
\usepackage{booktabs}
\usepackage{float}
\newcommand{\ra}[1]{\renewcommand{\arraystretch}{#1}}
\usepackage{caption}
\usepackage{subcaption}
\usepackage{hyperref}
\usepackage{xcolor}
\usepackage{cite}
\usepackage{wrapfig}
\usepackage[dvipsnames]{xcolor}

\newcommand{\Red}[1]{\textcolor{BrickRed}{\textbf{#1}}}

\begin{document}
\title{Automated Domain Question Mapping (DQM) with Educational Learning Materials}
\author{
Jiho Noh, Mukhesh Raghava Katragadda,  Dabae Lee  
\thanks{J. Noh is with the Cognition and Learning Design Lab, Department of Computer Science, Kennesaw State University, GA, 30060 USA (e-mail: jnoh3@kennesaw.edu)}
\thanks{M. R. Katragadda is with the Department of Computer Science, Kennesaw State University, GA, 30060 USA (e-mail: mkatrag1@students.kennesaw.edu)}
\thanks{D. Lee is with the School of Instructional Technology and Innovation, Kennesaw State University, GA, 30060 USA (e-mail: dlee159@kennesaw.edu)}
} 


\markboth{IEEE TRANSACTIONS ON LEARNING TECHNOLOGIES, VOL. xx, NO. x, JANUARY 20xx}%
{Jiho Noh \MakeLowercase{\textit{(et al.)}}: Automated Domain Question Mapping (DQM) with Educational Learning Materials }

\maketitle

\begin{abstract}

Concept maps have been widely utilized in education to depict knowledge structures and the interconnections between disciplinary concepts. Nonetheless, devising a computational method for automatically constructing a concept map from unstructured educational materials presents challenges due to the complexity and variability of educational content. We focus primarily on two challenges: (1) the lack of disciplinary concepts that are specifically designed for multi-level pedagogical purposes from low-order to high-order thinking, and (2) the limited availability of labeled data concerning disciplinary concepts and their interrelationships. To tackle these challenges, this research introduces an innovative approach for constructing Domain Question Maps (DQMs), rather than traditional concept maps. By formulating specific questions aligned with learning objectives, DQMs enhance knowledge representation and improve readiness for learner engagement. The findings indicate that the proposed method can effectively generate educational questions and discern hierarchical relationships among them, leading to structured question maps that facilitate personalized and adaptive learning in downstream applications.

\end{abstract}

\begin{IEEEkeywords}
Concept map, domain question map construction, large language models, concept extraction, question generation, AI in education
\end{IEEEkeywords}



\section{Introduction~\label{sec:introduction}}

\IEEEPARstart{T}{he} current education landscape is witnessing a significant increase in the utilization of machine learning and artificial intelligence (ML/AI) technologies, especially for online and hybrid learning, shifting towards student-centered learning approaches. This transformation necessitates an appropriate design and prompt implementation of personalized and adaptive learning \cite{locTransformingAmerican}. AI-powered educational applications, such as intelligent educational information systems, adaptive learning platforms, and intelligent tutoring systems, are being developed to provide personalized learning experiences for students \cite{Miao2021-bw,Tapalova2022-zm,Lee2025-qt}, catering to their individual needs and learning preferences, and supporting teachers in identifying gaps in individual students’ learning process. A commonly used framework for these approaches is \textit{Knowledge Tracing (KT)}, and one of its primary components is the domain model, which captures structural information of knowledge concepts and topics, such as \textit{concept maps}.

Concept maps~\cite{Reigeluth2009BuildingAC,Reigeluth2009InstructionalDesignTA} are graphical representations of knowledge structures that illustrate the relationships between various disciplinary concepts within a specific domain. The underlying theories of learning and cognition that support the use of concept maps are rooted in \textit{constructivism} and \textit{cognitivism}. Constructivism posits that learners actively construct their understanding of the world through experiences and interactions by integrating new incoming information with their existing knowledge~\cite{Driscoll1993PsychologyOL}. Cognitivism emphasizes the role of mental processes in learning, suggesting that learners organize and structure information in meaningful ways~\cite{ansari1968educational}. In this context, disciplinary concepts are defined as the fundamental building blocks of knowledge within a specific domain, and the map of these concepts can be used to explain the students' knowledge acquisition \cite{novak1984learning,wertsch1988vygotsky}.

Disciplinary concepts encompass the essential knowledge and principles in a particular subject domain, which students are expected to comprehend and apply. They are different from topics, keywords, and index terms, which are often used to categorize or label content. These concepts are generally more abstract and comprehensive than specific topics or keywords, which often refer to narrower subjects or specific instances within a broader domain. Disciplinary concepts are not just isolated pieces of information; instead, they are interrelated, forming a network of knowledge that helps learners understand the subject matter more comprehensively.

This particular definition of disciplinary concepts leads to several challenges in concept extraction and map construction. \ul{First, a concept can be defined and explained differently across various domains.} Hence, accurately identifying and extracting concepts from unstructured text requires a deep understanding of the subject matter and contextual knowledge. \ul{Second, the relationships between disciplinary concepts are often complex and multifaceted, and their representations in text are not evident or explicit.} These relationships can include hierarchical structures, causal connections, and associative links. Capturing these intricate relationships in a concept map is in and of itself a challenging task. \ul{Third, unlike keywords, the scarcity of labeled data for disciplinary concepts and their relationships makes it difficult to train supervised models for concept extraction and map construction.} This issue is especially prevalent in specialized domains where the available data may be scarce and unrepresentative.

In recent years, the rate of technological advancement has accelerated significantly, resulting in a substantial reduction in \textit{the half-life of knowledge}. Consequently, the conventional education system, which primarily focuses on disseminating knowledge, is no longer sufficient to meet the demands of contemporary education. Learners are now expected to acquire skills that enable them to adapt to the rapidly changing knowledge landscape. As such, the learning paradigm is shifting towards a more interactive and dynamic approach, where students are encouraged to engage actively with the context. Therefore, it is crucial to implement robust methodologies for automatically converting static learning materials into interactive learning-by-doing resources. The automatic construction of structured knowledge representations, such as concept maps, is a crucial task within this framework for developing personalized learning systems.

In this study, we further develop a method for constructing a \textbf{Domain Question Map (DQM)}, a map of questions rather than concepts. The rationale for this approach is to enhance the representation of the knowledge structure by formulating questions that are more specific and pertinent to the topics and knowledge components within a domain. For example, rather than using a general concept like ``creativity assessment'' in education, we can create a question like ``Which linguistic or visual aspects of students' responses can be computationally used to assess creativity?'' Creating such questions ensures they are more specifically aligned with the learning objectives, thereby creating and facilitating more closely aligned instruction and assessment.

We aim to address some of the aforementioned research gaps by proposing a novel approach to automatic DQM construction, which involves fine-tuning masked language models (MLMs) or utilizing large-scale causal language models (CLMs). Recent research has shown that the prior knowledge captured within LLMs can be effectively leveraged to automatically construct structured knowledge representations, such as knowledge graphs, ontologies, and concept maps~\cite{Funk2023-eo,Wehnert2024-vx}. We propose using the most advanced AI technologies to construct question maps in educational domains. Our methodology is founded on the hypothesis that AI technologies, particularly LLMs, can comprehend educational materials, generate questions, and identify the relationships among these questions to construct a practically valuable knowledge map.

Accordingly, we address the following research questions in this study:
\begin{itemize}
  \item RQ1. For the purpose of generating domain questions, how effectively can pre-trained language models be fine-tuned with general question-answering benchmark datasets?
  \item RQ2. Can the hierarchical outline of a textbook be an effective source for training a classification model to identify the relationships between domain questions?
  \item RQ3. Which graph operations can be utilized to enhance the quality of AI-generated domain questions?
\end{itemize}

\section{Background~\label{sec:background}}

\subsection{Developing a Knowledge Graph Framework for Educational Applications}
\noindent
In the domain of information processing, the term ``concept'' is often used interchangeably with ``keyword'' or ``topic.'' However, these terms hold distinct meanings, particularly within the educational context. Erickson has emphasized the transfer of learning, where the transfer occurs at the conceptual level of understanding~\cite{Erickson2002-xe,Erickson2008-wz}.  She presents the hierarchical structure of knowledge, progressing from the most abstract to the most concrete levels: \textit{concepts}, \textit{topics}, and \textit{facts}. In this context, learners generalize concepts to a theory and understand the relationships between concepts. Essentially, concepts are timeless, universal, and abstract ideas that allows for deeper understanding through various examples. Topics, in comparison, are specific subjects that often lead to fact-based knowledge. For example, ``Energy Conservation'' can be a concept based on the topic of ``Velocity.'' The term \textit{keyword} is more information retrieval-oriented and is often used to refer to specific terms or phrases that are used to categorize or label content, which can be extracted from the given text using natural language processing techniques.

The concept map is a widely used tool in education that visually represents the relationships between concepts, aiding various educational processes. Novak~\cite{novak1984learning} defined the nodes in a concept map as ``perceived regularities in events or objects designated by a label,'' and the links between nodes as the ``ideational relationship between two or more concepts.'' Educational researchers and practitioners, including instructional designers and curriculum developers, have long recognized the importance of mapping disciplinary concepts to facilitate learning~\cite{Reigeluth2009InstructionalDesignTA}. 

A large portion of the research on concept maps has focused on their use as \textit{Mindtools}, which are tools that computer systems use to engage learners in constructive and higher-order thinking about subjects they are studying~\cite{Jonassen1998computers, chu2010knowledge, hwang2011concept, hwang2014definition}. They serve as knowledge construction tools that students learn with, rather than from. For example, \textit{Stella} is a mindtool that allows students to create dynamic models of complex systems, enabling them to visualize and manipulate the relationships between different components of the system. CMapTools is another concept mapping software that allows users to create, share, and collaborate on concept maps. 
These tools exhibit certain limitations, including unidirectional construction of knowledge that needs manual input, as well as challenges in maintaining the maps as the knowledge evolves.

A Concept map serves as a domain model that captures the structure of knowledge in a specific subject area, allowing for personalized learning experiences~\cite{abdelrahman2023knowledge, shen2024survey}. Especially with the Knowledge Tracing (KT) framework, effectively representing the knowledge structure of a domain has been a long-standing research goal~\cite{Lu2022-tg, JiangUnknown-na, Park2024-dc, Gao2021-fv, Xia2023-ye, Panjaburee2010-tq}. Concept maps have also been used as an \textit{assessment strategy} to evaluate students' understanding and conceptions of a subject matter by examining students' knowledge state in comparison to the domain model~\cite{Burkhart2020-jj, Lin2019-di}.

\subsection{Technologies for Concept Map Construction}
\noindent
The body of previous research on the use of concept maps and their automated construction is extensive, showing a clear progression from traditional knowledge graph construction to modern AI-driven methods, with a particular emphasis on knowledge representation through the integration of large language models. We can find various knowledge engineering technologies that have been applied to education domains, including: (1) concept/relation extraction and analysis, (2) knowledge graph construction, and (3) generative AI for concept extraction and map construction.

FACE~\cite{Chau2021-ho} focused on concept extraction using a supervised machine learning model with a list of hand-crafted linguistic features from digital textbooks, including information such as Part-of-Speech (POS) tags, word frequency, word length, and the hierarchical structure of the textbook. WERECE~\cite{Huang2023-pn} proposed an unsupervised method of using word embeddings for disciplinary concept extraction. \cite{Wang2016-to} conceptualized the construction of concept maps as a joint optimization problem, integrating both key concept extraction and prerequisite relation extraction. More recently, \cite{Reales2024-hh} utilized LLMs to construct a domain knowledge graph from educational material guided by ontologies like DBPedia, which are then used to identify core concepts by applying PageRank.

The automatic construction of knowledge graphs is another primary focus of research in educational technology, and the efficacy of LLMs in this area has been demonstrated in various studies~\cite{Abu-Salih2024-ey, Su2020-ar, Ding2024-nc, Yang2024-bi, Kommineni2024-hd} along with language model-based technologies such as prompt engineering, retrieval-augmented generation (RAG), and GraphRAG. In common, LLMs provide prior knowledge that can be leveraged to extract disciplinary concepts and identify relationships between them, a method that is more effective than traditional methods that rely on hand-crafted rules or supervised learning with limited labeled data.

While there has been significant progress in concept extraction and knowledge graph construction, few studies have focused on extracting disciplinary concepts designed explicitly for pedagogical purposes, which are better aligned with Erickson's definition of concepts. 

\subsection{Automatic Question Generation for DQM Construction}
\noindent
One of the limitations in concept map construction is the weak definition of disciplinary concepts, which can lead to ambiguity and inconsistency in the representation of knowledge. Another challenge is the lack of data for disciplinary concepts and their relationships, which makes it challenging to train supervised models for concept extraction and map construction. As previously stated, we propose a novel approach to DQM construction using question generation (QG) technologies. This approach can produce questions that are more specific and pertinent to the targeted learning objectives within a domain. Furthermore, training a model to generate questions is advantageous due to the abundance of labeled data available for question-answering tasks, which can be used to fine-tune pre-trained language models (PLMs) for QG.

In this study, the QG task is not directly intended for assessment purposes, but rather for the DQM generation to capture domain concepts for educational and instructional purposes. There is a substantial body of research on QG for assessment purposes~\cite{Wang2022-zb,Nguyen2022-dy,Scaria2024-zb,Ming2025-rg}, which our model will be based on. However, the evaluation criteria for QG models in the context of DQM construction are different from those used for assessment purposes~\cite{Gorgun2024-wo,Wang2023-cg, Hwang2023-fs}.

\section{Methodology~\label{sec:methodology}}

\subsection{Question Generation (QG)}
\noindent
As mentioned earlier, publicly available benchmark datasets for question-answering (QA) tasks are abundant. The motivation underlying this study is to leverage existing QA datasets for generating educational questions that align better more closely with learning objectives than with knowledge concepts within a domain.

\subsubsection{QA Datasets}
\noindent
For QG, we utilized two datasets: the KhanQ and SQuAD datasets. While other datasets, such as the MammothQA dataset, were considered, it was excluded from this study due to its primary focus on assessment purposes and the lack of sufficient contextual information required for effective question derivation. The KhanQ dataset~\cite{gong2022khanq} comprises 1,034 questions in the STEM fields (i.e., biology, chemistry, physics, environmental science, and electrical engineering) aiming to provide an in-depth understanding of the subjects taught in \textit{Khan Academy}, a well-known online education platform. The SQuAD (version 2.0) dataset~\cite{Rajpurkar2016-ga} is a commonly used benchmark dataset for question-answering tasks, comprising 142,192 question-answer pairs that covers a wide range of topics (e.g., historical events, scientific concepts, notable figures, geographical locations, institutions, and cultural subjects) from \textit{Wikipedia} articles. This version includes both answerable and unanswerable questions; however, for our task, we used only the answerable questions. Table~\ref{tbl:ds-stats} shows the descriptive statistics of the datasets.

\begin{table*}[htbp]
  \centering
  \caption{Descriptive statistics of the datasets used for training question generation models.}\label{tbl:ds-stats}
  \begin{tabular}{@{}lrrrrl@{}}
    \toprule
    & \multicolumn{2}{c}{Number of Samples} & \multicolumn{2}{c}{Avg. Length (words)} & \\
    \cmidrule{2-3} \cmidrule{4-5}
    Dataset & Context & Question & Context & Question & Most Common Interrogative Words (Uni/Bigrams)\\
    \midrule
    KhanQ & 723 & 723 & 83.8$\pm$64.1 & 10.6$\pm$4.7  &
    \begin{tabular}[t]{@{}l@{}}
      -- Why (21.1\%), How (19.5\%), What (11.0\%), Does (9.2\%), Is (8.6\%)\\
      -- Why does (3.9\%), Why is (3.3\%), How does (3.2\%), Does the (3.2\%), What is (2.8\%)
    \end{tabular} \\
    SQuAD & 19,035 & 86,821 & 116.6$\pm$49.7 & 10.1$\pm$3.6 &
    \begin{tabular}[t]{@{}l@{}}
      -- What (42.9\%), Who (9.3\%), How (9.3\%), When (6.2\%), In (5.0\%)\\
      -- What is (8.4\%), What was (5.4\%), How many (4.9\%), When did (3.1\%), In what (2.9\%)
    \end{tabular} \\
    \bottomrule
  \end{tabular}
\end{table*}

\subsubsection{Fine-tuning Pretrained Language Models}
\noindent
We fine-tuned four pre-trained language models (PLMs) for the QG task, including both encoder-decoder and decoder-only architectures. We also compare the performance of these models with the commercial instruction-tuned large language models (LLMs). The fine-tuning results of the models along with the model specifications can be found in Table~\ref{tbl:qg-eval}.

The model training configurations vary depending on the model architecture and size. For the encoder-decoder models (e.g., Pegasus), we used a batch size of 8, a learning rate of 5e-5, and a maximum sequence length of 768 tokens for the input context and 256 tokens for the output question. The model was trained for six epochs using the AdamW optimizer with a weight decay factor of 0.1 and a learning scheduler (cosine annealing) with warm-up steps. All models were implemented using the PyTorch framework and the Hugging Face Transformer models and trained on NVIDIA H100 GPUs.

\subsubsection{Evaluation Methods}
\noindent
Various evaluation metrics exist for text generation tasks~\cite{Celikyilmaz2020-dq}. We used four metrics in this study: two untrained automatic metrics (i.e., \texttt{BLEU} and \texttt{ROUGE}) and two machine-learned metrics (i.e., \texttt{BERTScore} and \texttt{BLEURT}). Both \texttt{BLEU}~\cite{Papineni_undated-yr} and \texttt{ROUGE}~\cite{Lin2004-bu} are widely used n-gram overlap metrics that measure the degree of overlap between the generated text and the reference text. \texttt{BERTScore}~\cite{Zhang2019-sf} is a metric that leverages pre-trained contextual embeddings from BERT~\cite{Devlin2019-yt} to measure the similarity between the generated text and the reference text. \texttt{BLEURT}~\cite{Sellam2020-pe} is another learned evaluation metric that uses a fine-tuned BERT model, optimized based on various intrinsic evaluation metrics (e.g., \texttt{BLEU} and \texttt{ROUGE}) and human references. 

\subsection{Specificity Classification of Generated Questions~\label{sec:spec-classification}}
\noindent
From an epistemological perspective, several definitions address the interrelationships among knowledge concepts, including but not limited to: (1) specialization, (2) composition, (3) commonality, and (4) dependency. This study primarily highlights the first two types of relationships, which are prevalent in taxonomies or ontologies. These relationships are advantageous for structuring knowledge hierarchies in tasks such as \textit{learning path suggestion} and \textit{knowledge graph construction}. 

The study's focus on specialization and composition in knowledge hierarchies is implemented through the specificity-relationship prediction model, which identifies the type of relationship between generated questions. Given a pair of questions ($q_i$, $q_j$) and their corresponding contexts ($c_i$, $c_j$), the model ($\Phi$) predicts the relationship type $r_{ij}$, whether one question is more general or specific than the other, or if no relationship exists, which can be formally defined as follows:
\begin{equation}
  \Phi(q_i, q_j, c_i, c_j) = r_{ij} \in \{\textit{general}, \textit{specific}, \textit{other}\}
\end{equation}

The specificity predictions can be used to create a directed edge between $q_i$ and $q_j$ if the relationship is either \textit{general} or \textit{specific}. No edge is created if the relationship is \textit{other}.

In this study, we used the confidence score ($\eta_{ij} \in [0, 1]$) to indicate the certainty that a pair of questions belongs to either the \textit{general} or \textit{specific} classes. Although using a softmax output directly as a confidence score can be misleading due to the overconfidence issue of neural networks~\cite{Hendrycks2016ABF}, it remains useful when used as a proxy for relative confidence between pairs of questions. The specificity relationship score is defined as follows, which is later used in the graph post-processing step (Eq. \ref{eq:edge-weight}).
\begin{equation}
  \eta_{ij} = 1 - softmax(r_{ij} = \textit{other})
\end{equation}

\subsubsection{Data Preparation~\label{sec:rel-cls-data}}
\noindent
To prepare for training a hierarchical relationship prediction model, we compiled a set of context pairs from typical college-level textbooks and assigned relationship labels to them. The relationship type is derived by comparing the section numbers from which the text chunks are taken. The section number of a text chunk is identified by locating the nearest preceding section heading. A section ID follows this format: `\mbox{SECTION\#\#\#\#\#\#\#\#\#\#}'. Each two-digit segment indicates a section number, starting from the highest level on the left, prefixed by `SECTION'. For example, `SECTION0104030000' is for Section 1.4.3. We can derive the relationship type between sections `010400` and `010403` as a direct \textit{parent-child} relationship, such that the pair of contexts from these sections, in the specified order, would be labeled as `\textit{general}'.

The data preparation process involves the following steps:

\begin{enumerate}[label=(\alph*)]
  \item Convert a textbook in PDF format into a Markdown format text file; we used the \textit{Mistral-AI} OCR tool~\footnote{\url{https://mistral.ai/news/mistral-ocr}} for this purpose. 
  \item Split the textbook into discrete text segments, ensuring that each segment remains confined within a single section. Each text segment comprises a paragraph or multiple consecutive paragraphs that are semantically coherent. For this task, we employed the \textit{LangChain Semantic Chunker}~\footnote{\url{https://www.langchain.com/}}.
  \item Compile all possible pairs of text segments (i.e., contexts) that are in the \textit{parent-child} relationship based on their hierarchical position information from the section numbers. 
  \item Swap the order of the text segments in each pair to create \textit{child-parent} pairs.
  \item To create \textit{other} pairs, randomly sample the same number of examples as the \textit{parent-child} pairs from all possible pairs that do not have a \textit{parent-child} relationship or \textit{child-parent} relationship.
\end{enumerate}

\subsubsection{Evaluation Methods}
\noindent
We evaluated the specificity classification model using standard multi-class classification metrics, including precision, recall, and F1-score. The model's performance was assessed on a held-out test set that was not used during training. Additionally, we analyzed the confusion matrix to understand the types of errors made by the model and to identify areas for improvement.

\subsection{Graph Post-Processing}
\noindent
A standard college-level textbook typically contains between 300 and 1,000 pages. Assuming each page can be divided into 3 to 5 semantically coherent text segments, we expect to generate between 900 and 5,000 questions from a single textbook. This volume of questions is not suitable for practical use in a DQM, as it would hinder users' ability to navigate the map or comprehend the topics within it effectively. Consequently, it is necessary to employ graph post-processing techniques to refine and reduce the size of the generated question graph.

\subsubsection{Reducing the Number of Question Nodes}
\noindent
Given a specified target number of questions, denoted as $n$ (e.g., 300), and an initial set of $N$ generated questions, we iteratively merge the most semantically similar pairs of questions a total of $(N-n)$ times until the desired number is achieved, such as in the hierarchical clustering process. To find the similarity of two questions, we employ a Transformer-based sentence encoding model, Sentence-BERT~\cite{Reimers2019-yq}, to convert each question $q$ along with its source context $c$ into a fixed-length vector representation. The cosine similarity between two question vectors is then calculated to determine their semantic similarity.

\begin{equation}
\xi_{ij} = cossim\left(\Pi(q_i \Vert c_i), \Pi(q_j \Vert c_j)\right)
\end{equation}

To merge two question nodes in a graph, we can create a new question that encompasses both of them. This method is non-trivial and requires careful consideration to ensure that the new question accurately reflects the content of both original questions. We leave this approach for future work. Instead, we opt for selecting the more \textit{general} question over the other, utilizing the pre-trained specificity classification model outlined in Section~\ref{sec:spec-classification}. The assumption is that a more \textit{general} question is likely to encompass the knowledge of two paired questions, thereby serving as a more representative question for the relevant topic.

\subsubsection{Connecting the Question Nodes~\label{sec:connect-nodes}}
\noindent
The desired characteristics of the DQM include the ability to identify learning paths and sub-topics within a domain. To achieve this, it is essential to ensure that the graph exhibits a certain level of connectivity such that (1) all nodes are reachable from any other node without cycles and (2) the graph avoids being overly dense.

We use the \textit{Minimum Spanning Tree (MST)}, a powerful graph sparsification technique that removes cycles from weighted graphs. The MST algorithm removes less weighted edges, preserving node reachability by minimizing the total edge weight. The MST algorithm is suitable for our task because it enables us to create a tree-like structure in the question graph, which is desirable for educational purposes.

When the edge weights are considered as the cost of traversing the edge, the MST algorithm can be used to find the most efficient path between two nodes in a graph. In our case, the weights ($w$) measure the hierarchical relationship ($\eta$) and the cohesiveness ($\xi$) of the questions; hence, preserving the strongest relationships between questions is desirable for maintaining the integrity of the DQM. 
\begin{equation}
  w_{ij} = \lambda \eta_{ij} + (1 - \lambda) \xi_{ij}, \quad \lambda \in [0, 1]
  \label{eq:edge-weight}
\end{equation}

To achieve the same effect as the MST algorithm, we multiply the edge weights by a factor of $-1$ to convert the problem into a \textit{Maximum Spanning Tree (MaxST)} problem, and then apply Kruskal's~\cite{Kruskal1956-ps} algorithm to find the MaxST.

The following steps outline the graph post-processing procedure:
\begin{enumerate}[label=(\alph*)]
  \item Add nodes of all the generated questions to the graph
  \item Iteratively merge the most semantically similar pairs of questions until the desired number of questions is reached.
  \item Create a complete graph by adding edges between every pair of questions, where the edge weight is the relationship score between the questions (See Eq.~\ref{eq:edge-weight}).
  \item Prune the graph by applying the Maximum Spanning Tree (MaxST) algorithm to remove cycles and preserve the most significant relationships between questions.
\end{enumerate}

\vspace{1em}
\section{Results~\label{sec:results}}

\subsection{Evaluation of Question Generation Models [RQ1]}
\noindent
Various models are trained and evaluated on the KhanQ and SQuAD datasets for the QG task. Table~\ref{tbl:qg-eval} shows the evaluation results of the models on both datasets. The models being compared encompass both fine-tuned pre-trained language models, including both the encoder-decoder and decoder-only architectures, as well as commercially available instruction-tuned large language models (LLMs) evaluated in zero-shot and few-shot settings. The results indicate that the performance of the models varies significantly depending on the dataset and the model architecture. With all the models, the performance on the KhanQ dataset is generally lower than that on the SQuAD dataset, which can be attributed to the smaller size of the KhanQ dataset. Among the PLMs, \ul{the encoder-decoder models outperform the decoder-only model} on both datasets across all evaluation metrics. However, we could not conclude that it is the primary factor for the performance difference, as the model sizes also vary significantly. \ul{\textit{GPT-o4-mini} model performed competitively} with the fine-tuned PLMs, especially in the few-shot setting.

\begin{table*}[htbp]
  \centering
  \caption{Evaluation results of question generation models on the KhanQ and SQuAD datasets. The best scores for each metrics are highlighted in \textbf{bold}. Evaluation metrics used are BLEU (BLU), ROUGE-L (RGL), BLEURT (BLT) and BERTScore (BRT). \label{tbl:qg-eval}}
  \begin{tabular}{@{}lrrcrrrcrrrr@{}}
    \toprule
    & & & \multicolumn{4}{c}{KhanQ} & & \multicolumn{4}{c}{SQuAD} \\
    \cmidrule{4-7} \cmidrule{9-12}
    & Architecture & Parameters & BLU & RGL & BLT & BRT && BLU & RGL & BLT & BRT \\
    \midrule
    BART-large & Enc-Dec & 406 M & 0.005 & 0.153 & 0.341 & 0.869 && \textbf{0.281} & 0.411 & 0.434 & 0.915 \\
    Pegasus-large & Enc-Dec & 568 M & 0.003 & 0.156 & 0.347 & 0.869 && 0.217 & 0.480 & 0.426 & 0.906 \\ 
    T5-Large-Squad-QG & Enc-Dec & 770 M & 0.009 & \textbf{0.193} & \textbf{0.357} & \textbf{0.879} && 0.263 & \textbf{0.534} & \textbf{0.538} & \textbf{0.937} \\
    GPT-2 & Dec-only & 355 M & 0.013 & 0.147 & 0.324 & 0.850 && 0.192 & 0.389 & 0.401 & 0.895 \\
    \midrule
    GPT-o4-mini (zero-shot) & Dec-only & 8 B & 0.014 & 0.148 & 0.278 & 0.853 && 0.113 & 0.387 & 0.453 & 0.908 \\
    GPT-o4-mini (few-shot) & Dec-only & 8 B &\textbf{0.023} & 0.157 & 0.328 & 0.867 && 0.127 & 0.411 & 0.467 & 0.911 \\
    \bottomrule
  \end{tabular}
\end{table*}

We utilized one of the fine-tuned models (i.e., T5-Large-Squad-QG) to generate questions from the SQuAD dataset, providing examples of high-, medium-, and low-scored questions. The examples in Table~\ref{tbl:qg-examples} illustrate the varying quality of the generated questions, highlighting the challenges in generating high-quality questions that are both thought-provoking and relevant to the given context.

\begin{table*}[tb]
  \centering
  \footnotesize
  \ra{1.1}
  \caption{Examples of questions generated by the fine-tuned T5-Large-Squad-QG model, derived from the SQuAD dataset, and categorized as high, medium, and low quality based on the similarity scores.}\label{tbl:qg-examples}
  \begin{tabular}[t]{@{}p{.5\textwidth}p{.5\textwidth}@{}}
    \toprule
    \textbf{Source Text} & \textbf{Questions and Similarity Score}\\
    \midrule
    Southern California, often abbreviated SoCal, is a geographic and cultural region that generally comprises California's southernmost 10 counties. The region is traditionally described as "eight counties", based on demographics and economic ties: Imperial, Los Angeles, Orange, Riverside, San Bernardino, San Diego, Santa Barbara, and Ventura. 
    & 
    \begin{tabular}[t]{@{}p{.47\textwidth}@{}}
      \textbf{Generated:} What is the abbreviation for Southern California?\\
      \textbf{Ground Truth:} What is Southern California often abbreviated as?\\
      \textbf{Scores:} \\
      0.077 (BLEU) 0.571 (ROUGE-L) 0.684 (BLEURT) 0.927 (BERTScore)
    \end{tabular} \\ \midrule
    `Strong' and `weak' acids shouldn't be confused with higher or lower pH. What defines a strong acid is that it disassociates completely in water. What defines a weak acid is that it doesn't disassociate completely in water. For example: If a strong acid is in really low concentration (0,00001M) in one solution, and a weak acid is in relatively high concentration (1M) in another, the H+ concentration would probably be higher in the weak acid solution, and it would therefore have lower pH(despite being the weaker acid.)
    & 
    \begin{tabular}[t]{@{}p{.48\textwidth}@{}}
      \textbf{Generated:} What should 'strong' and 'weak' acids not be confused with higher or lower pH?\\
      \textbf{Ground Truth:} Can the strength of a base or acid be determined only by its starting pH? \\
      \textbf{Scores:}\\
      0.018 (BLEU) 0.207 (ROUGE-L) 0.345 (BLEURT) 0.879 (BERTScore)
    \end{tabular} \\ \midrule
    Superfluid means the substance is at the point between liquid and gas (equilibrium) at very high temperature and pressure. helium becomes a superfluid at extremely low temperatures. superfluid is strange in the way that it seems to have a viscosity of zero, it can flow up the sides of a container, and many other creepy things. &
    \begin{tabular}[t]{@{}p{.48\textwidth}@{}}
      \textbf{Generated:} What is superfluid?\\
      \textbf{Ground Truth:} Does applying more heat not change the average kinetic energy? \\
      \textbf{Scores:}\\
      0.000 (BLEU) 0.000 (ROUGE-L) 0.213 (BLEURT) 0.855 (BERTScore)
    \end{tabular} \\
    \bottomrule
  \end{tabular}
\end{table*}

\subsection{Relationship Classification Performance [RQ2]}
\noindent
For training a relationship classification model, we used a textbook in the computer science domain, titled ``\textit{An Introduction to Information Retrieval}''~\footnote{\url{https://nlp.stanford.edu/IR-book/pdf/irbookonlinereading.pdf}} by Manning et al. As described in Section~\ref{sec:rel-cls-data}, we converted the textbook from PDF to Markdown format, split the text into semantically coherent chunks, and created pairs of text chunks with relationship labels based on their hierarchical structure. The statistics of the processed textbook are summarized in Table~\ref{tab:irbook-stats}.

\begin{table}[H]
  \caption{Text statistics of the processed textbook (IR Book)}\label{tab:irbook-stats}
  \begin{center}
    \begin{tabular}{@{}lr@{}}
      \toprule
      \textbf{Description} & \textbf{Value} \\
      \midrule
      \# of chunks & 1,071 \\
      \# of pair examples  & 7,275 \\
      \# of chapters / sections & 21 / 121\\
      depth (avg / max) & 2.7 / 4 \\
      \bottomrule
    \end{tabular}
  \end{center}
\end{table}

GPT-2 Medium was used for the relationship classification task. The model was trained using pairs of arbitrary chunks along with their relationship labels. The model was trained on the cross-entropy loss. The model achieved a macro average F1-score of 90\% on the test set, indicating that it effectively learned to classify the relationship types (Table~\ref{tab:rel-cls-results}). The model performed worst on the \textit{other} class, which is likely due to the fact that the \textit{other} class is the most diverse and ambiguous class.

\begin{table}[htbp]
  \caption{Specificity relationship classification results.}\label{tab:rel-cls-results}
  \begin{center}
    \begin{tabular}{@{}lrrr@{}}
    \toprule
       & \textbf{Precision} & \textbf{Recall} & \textbf{F1-Score} \\
    \midrule
      \textit{General} & 0.941 & 0.970 & 0.955 \\
      \textit{Specific} & 0.882 & 0.909  & 0.896 \\
      \textit{Other} & 0.871 & 0.818 & 0.844 \\
      \midrule
      Average & 0.899 & 0.899 & 0.899 \\
    \bottomrule
    \end{tabular}
  \end{center}
\end{table}

\subsection{DQM Construction [RQ3]}
\begin{figure*}[ht]
  \centering
  \begin{subfigure}[b]{0.3\textwidth}
    \centering
    \includegraphics[width=\textwidth]{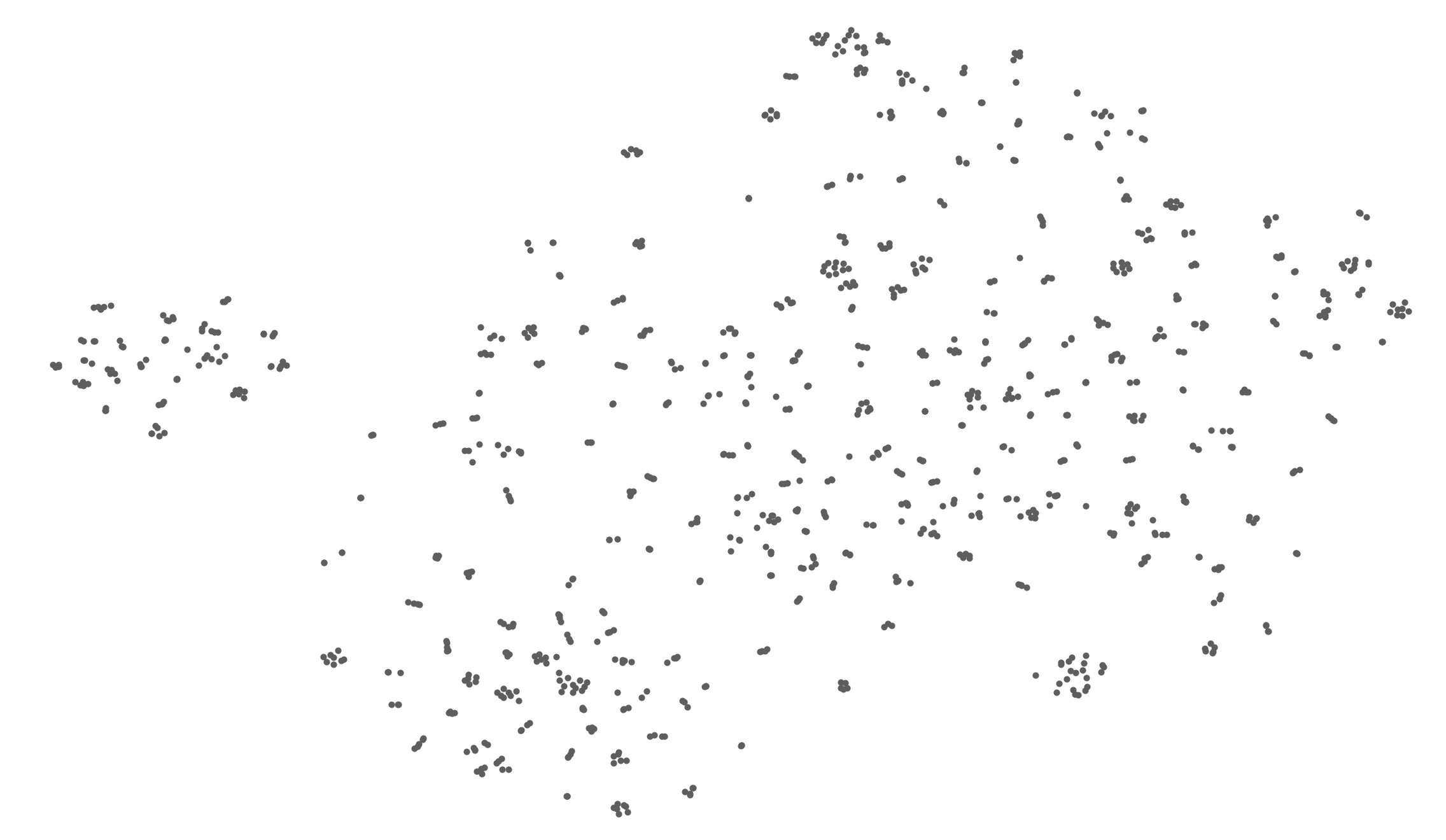}%
    \caption{Question embeddings\label{fig:pp-tsne}}
  \end{subfigure}
  \hfill
  \begin{subfigure}[b]{0.3\textwidth}
    \centering
    \includegraphics[width=\textwidth]{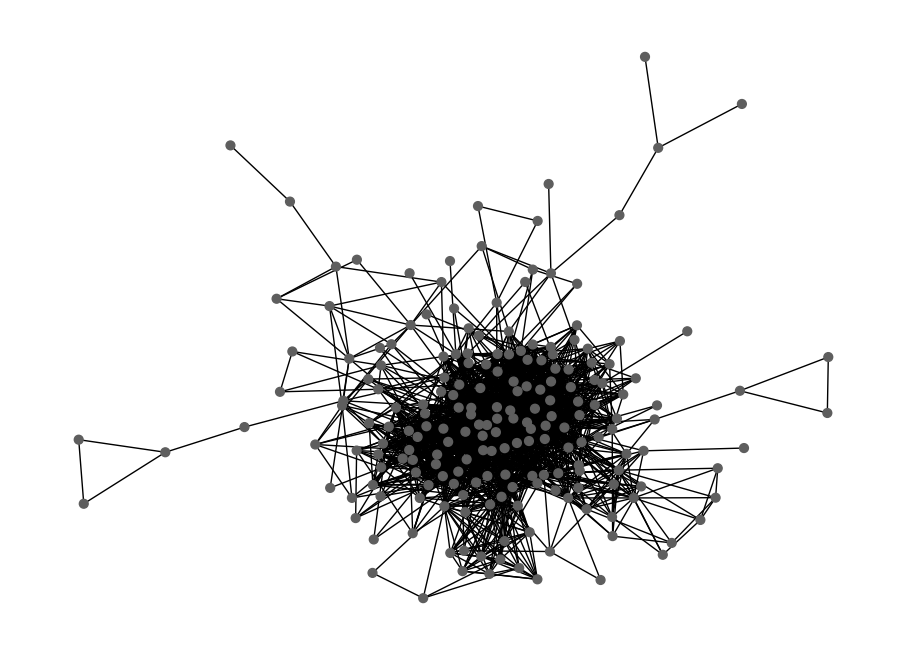}%
    \caption{Edge weight threshold selection~\label{fig:pp-edge-threshold}}
  \end{subfigure}
  \hfill
  \begin{subfigure}[b]{0.3\textwidth}
    \centering
    \includegraphics[width=.9\textwidth]{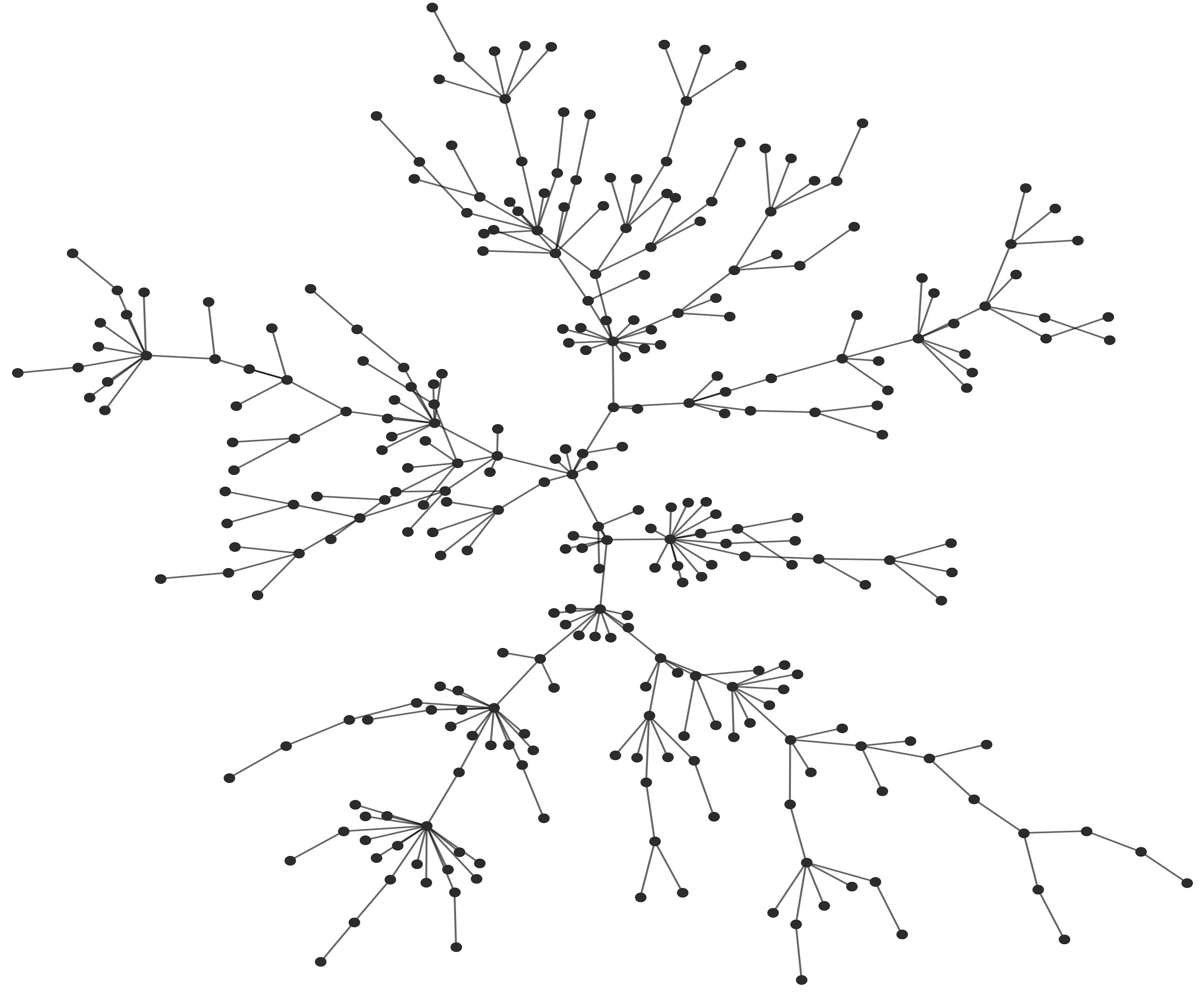}%
    \caption{Pruning of the graph using Minimum Spanning Tree (MST) by setting $\lambda=0.3$~\label{fig:pp-mst}}
  \end{subfigure}
  \caption{Results of graph post-processing. Figure (a) shows the t-SNE projection of question embeddings using Sentence-BERT, figure (b) shows the result of edge selection with a weight threshold ($\tau=0.7$), and figure (c) shows the final graph after pruning the graph using the MST algorithm. }
  \label{fig:graph-post-processing}
\end{figure*}

\noindent
For constructing a DQM using the same textbook, we generated 1,071 questions from the text chunks using the fine-tuned T5-Large-Squad-QG. The generated questions were then used to create a directed graph, where each question is a node, and the edges represent the relationships between questions. We utilized two measures that capture the relationships between questions: (1) the specificity relationship score ($\eta$) derived from the specificity classification model, and (2) the semantic similarity score ($\xi$) obtained using Sentence-BERT. The Figure~\ref{fig:graph-post-processing} illustrates the results of the graph post-processing steps. Figure~\ref{fig:pp-tsne} shows the t-SNE projection of question embeddings, which reveals some of the semantic clusters of questions. Figure~\ref{fig:pp-edge-threshold} shows the result of edge selection with a weight threshold ($\tau$=0.7) which still contains cycles and dense connections that prevents effective navigation of the graph. Finally, Figure~\ref{fig:pp-mst} shows the final graph after pruning using the MST algorithm, which ensures that all nodes are reachable from any other node without cycles.

\begin{figure*}[htbp]
  \begin{minipage}{0.35\textwidth}
    \centering
    \includegraphics[width=0.95\textwidth]{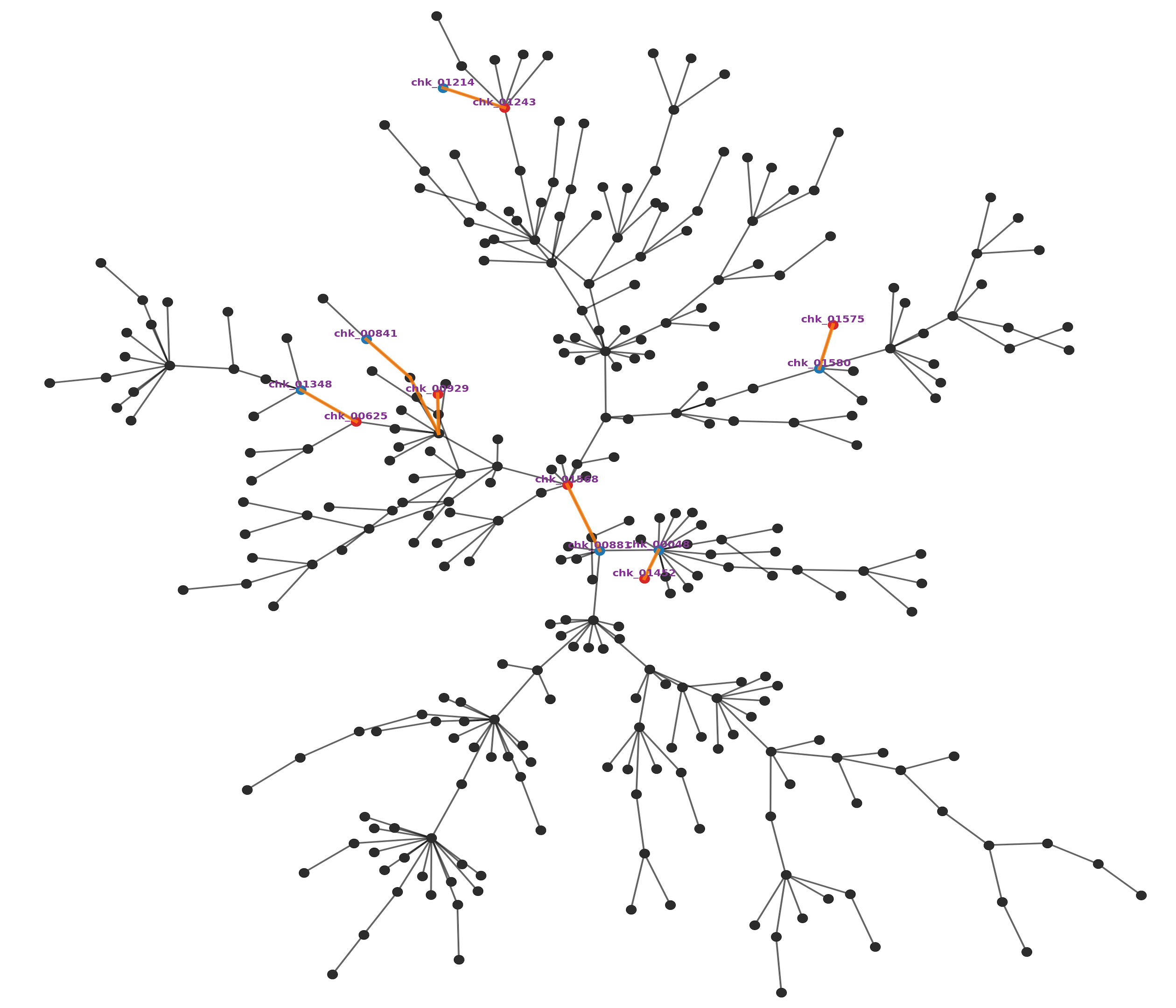}
  \end{minipage}
  \begin{minipage}{0.65\textwidth}
    \footnotesize
    \begin{itemize}
      \setlength\itemsep{0em}
      \item \Red{\texttt{chk\_00048}} What is the term for a query that is periodically executed on a collection to which new documents are incrementally added over time?
      \item \Red{\texttt{chk\_01452}} What is a common heuristic approach that web crawlers use to insert delays between successive fetch requests to the same host, typically an order of magnitude larger than the previous fetch time?\\[.3em]
      \item \Red{\texttt{chk\_00881}} What is the primary case of a statistical ranked retrieval model that naturally supports structured query operators?
      \item \Red{\texttt{chk\_01568}} What was the relevance feedback mechanism introduced in and popularized by Salton's SMART system around 1970?\\[.3em]
      \item \Red{\texttt{chk\_01214}} Which measures are best suited to compare the quality of the two clusterings?
      \item \Red{\texttt{chk\_01243}} What is the number of bits needed to transmit class memberships assuming cluster memberships are known?
    \end{itemize}
  \end{minipage}
  \caption{Sample paths over the final question graph after applying the MST algorithm. The question samples show the specificity relationship between the questions along the path.}
  \label{fig:graph-path}
\end{figure*}

As a part of the qualitative analysis of the final DQM, we randomly selected a path from the graph and evaluated the quality of the questions along that path. The selected path is shown in Figure~\ref{fig:graph-path}. The questions along the path are relevant to the topic of spelling correction in information retrieval, and they cover various aspects of the topic, including phonetic matching algorithms, context-sensitive spelling correction, threshold-based approaches, and feedback terms in retrieval quality. The questions are clear and concise, and they are likely to stimulate critical thinking and discussion among students. 

\section{Discussion~\label{sec:discussion}}

\noindent
In this study, we have explored various aspects of generating and organizing domain-specific questions to construct a DQM for educational purposes. We have tested several pre-trained language models and their performance. The results show that small-to-medium-sized PLMs can be effectively fine-tuned for the QG task using existing QA datasets. Nonetheless, the quality of the generated questions remains contingent upon the training data and the model used. The relationship classification shows robust performance, which can be trained using the hierarchical structure of textbooks. Additionally, the MST-based graph post-processing, employing semantic similarity and specificity relationship scores, helped shape the question graph into a more navigable structure by reducing the number of questions and improving the map's connectivity.

Several avenues for future research and development can be proposed. As mentioned in the methodology section, we should aim to develop methods for creating complex yet inclusive questions by synthesizing multiple similar questions. This advancement will enhance the depth and breadth of the question maps, providing a more comprehensive representation of the subject matter. Exploring different approaches to ingest various sources of teaching materials will be crucial. This includes integrating textbooks, lecture notes, and multimedia content to enrich the question maps and ensure they are reflective of diverse educational resources. We also plan to develop interactive tools that allow educators and students to engage dynamically with the question maps. These tools will facilitate real-time feedback and adaptation, enhancing the learning experience. Conducting field tests to assess the quality and effectiveness of the generated question maps in real educational settings will be essential. This will involve gathering feedback from educators and students to refine the models and improve the relevance and clarity of the questions.


\section{Acknowledgments}
\noindent
This work was supported in part by the Interdisciplinary Seed Grants from Kennesaw State University and the Computing Resources Research Grant from Lambda Cloud. The source code for this study is available at \href{https://github.com/YesNLP/Automated-Domain-Question-Map-Construction}{\textcolor{blue}{GitHub}}.

\bibliographystyle{IEEEtran} 
\bibliography{mybib}



%
%
%

\end{document}